\documentclass[12pt]{article}

\usepackage{sbc-template}
\usepackage{graphicx,url}
\usepackage[utf8]{inputenc}
\usepackage{changepage}

\title{Comparing Computational Architectures \\ for Automated Journalism}

\author{Yan V. Sym\inst{1, [0000-0002-0401-0586]}, João Gabriel M. Campos\inst{1, [0000-0002-1106-3694]},\\
Marcos M. José \inst{1, [0000-0003-4663-4386]}, Fabio G. Cozman \inst{1, [0000-0003-4077-4935]}}

\address{Escola Politécnica, Universidade de São Paulo, São Paulo, Brazil}

\begin{document} 
 
\maketitle

\begin{abstract}
The majority of NLG systems have been designed following either a template-based or a pipeline-based architecture. Recent neural models for data-to-text generation have been proposed with an end-to-end deep learning flavor, which handles non-linguistic input in natural language without explicit intermediary representations. This study compares the most often employed methods for generating Brazilian Portuguese texts from structured data. Results suggest that explicit intermediate steps in the generation process produce better texts than the ones generated by neural end-to-end architectures, avoiding data hallucination while better generalizing  to unseen inputs. Code and corpus are publicly available. \footnote{\url{https://github.com/C4AI/blab-reporter}}
\end{abstract}


\section{Introduction}\label{sec:introduction}

Natural Language Generation (NLG) is a subfield at the intersection of linguistics, computer science, and artificial intelligence, concerned with generating readable, coherent and meaningful explanatory text or speech so as to describe non-linguistic input data \cite{Ehud}. NLG is often viewed as complementary to Natural Language Understanding (NLU) and part of  Natural Language Processing (NLP). Whereas in NLU the goal is to understand input sentences to produce machine representations, in NLG the system must make decisions about how to transform representations into meaningful words and phrases \cite{liddy2001natural}.

Multiple successful examples of data-to-text systems can be found in weather forecasting \cite{sripada2004lessons}, financial and analytical reporting, industrial monitoring \cite{kim2020design} and conversational agents. Amongst NLG applications, robot-journalism is one of the most prominent endeavors thanks to the abundance of structured data streams available today, thus allowing automated systems to report recurring material with high-fidelity and lexical variation \cite{graefe2016guide}.

Traditionally, most data-to-text applications have been designed in a modular fashion as this facilitates reuse in different domains; going directly from input to output with rules has been simply too complex \cite{gatt2018survey}. In such systems, non-linguistic input data is converted into natural language through several explicit intermediate transformations and sequential tasks related to content selection, sentence planning and linguistic realization \cite{ferreira2019neural}. The two most frequently used automated journalism architectures are the template-based approach, which is application-dependent and lacks generalization capabilities due to its rule-based nature, and the pipeline-based approach, which embodies linguistic insights to convert data to text by applying a series of sequential steps.

The emergence of neural-based NLG systems in the recent years has changed the field: provided there is enough labeled data for training a machine learning model, learning a direct mapping from structured input to textual output has become reality \cite{li2017deep}. This has led to the recent development of deep learning end-to-end models, which directly learn input-output mappings and rely far less on explicit intermediary representations and linguistic insights.

Even though it is technically feasible to use neural end-to-end methods in real world applications, this does not necessarily mean that they are superior to rule-based approaches in every scenario. Recent empirical studies have demonstrated that a combination of template and pipeline systems produce texts that are more appropriate than the neural-based approaches, which frequently hallucinate content unsupported by the semantic input \cite{ferreira2019neural}. For the particular task of automated journalism, reporting inaccurate data would seriously undermine a robot’s credibility and could have serious implications on sensitive domains, such as environmental reports. A modular model also has the advantage of allowing for auditing, while neural end-to-end approaches behave as black-boxes \cite{campos2020towards}.

In this paper, we compare the three most frequently used architectures for automated journalism -- template-based, pipeline-based and end-to-end neural models -- using a common domain, the Blue Amazon. With an offshore area of 3.6 million square kilometers along the Brazilian coast, the Blue Amazon is Brazil's exclusive economic zone (EEZ); it is a oceanic region brimming with marine species and energy resources \cite{thompson2015blue}. 
Ocean monitoring, climate change and environmental sustainability are 
promising fields for automated journalism applications.
The oceans are severely damaged environments, and if current trends continues, there will be disastrous consequences for the planet as it is essential to halt climate change, fostering economic growth and preserving biodiversity \cite{e2022ocean}.
Although connecting with public audiences in an approachable way typically requires coverage by trained human journalists, accurate and low latency information reports can be very helpful. There is a vast and ever-growing body of information about the oceans; clearly, society can benefit from a robot journalism system. To address this issue, we created our robot journalism application which combines different NLG approaches to generate daily reports about the Blue Amazon and publish them on Twitter. \footnote{\url{https://twitter.com/BLAB_Reporter}}

A corpus of verbalizations of non-linguistic data in Brazilian Portuguese was created based on syntactical and lexical patterning abstracted from data collected from publicly available sources. Intermediate representations were annotated for each entry in order to develop our corpus. A combination of automatic and human evaluation together with a qualitative analysis was then carried out to measure the fluency, semantics and lexical variety of the generated texts.

This main contributions of this work are the construction of a publicly available Brazilian Portuguese NLG dataset, a comparison between the three most frequently used automated journalism architectures and an application which combines different approaches to publish daily reports about the Blue Amazon on Twitter. In Section \ref{sec:data}, we present our Blue Amazon dataset for automated journalism, and in Section \ref{sec:template} we discuss our approach in building a template-based architecture. Also, in Section \ref{sec:pipeline}, we present and discuss our pipeline architecture with six sequential modules. In Section 
\ref{sec:endtoend}, we discuss the end-to-end architecture and utilize it by training four different neural networks to generate the output text. In Section \ref{sec:results} we present the main results of this work and in Section \ref{sec:discussion} we discuss the results by providing some qualitative analysis. Finally, we conclude in Section \ref{sec:conclusion}.

\section{Non-linguistic Data about the Blue Amazon}\label{sec:data}

\begin{figure}[t]
\centering
\includegraphics[width=.600\textwidth]{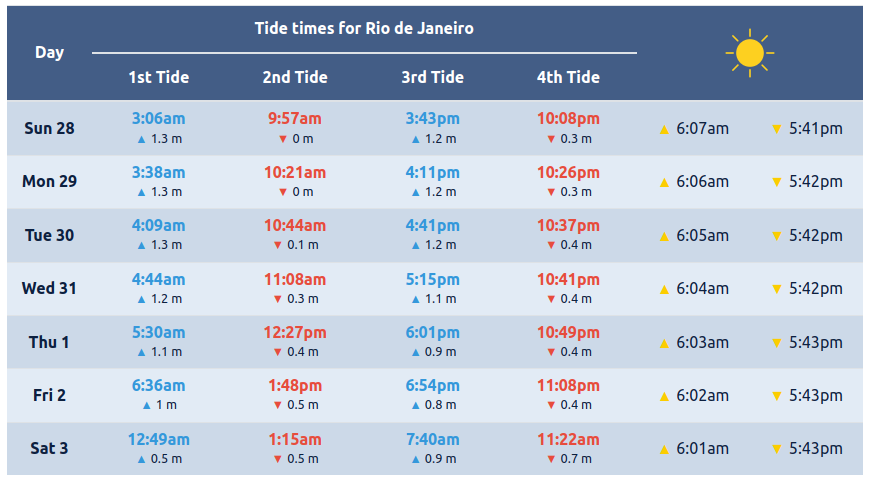}\hspace*{1mm}\includegraphics[width=.378\textwidth]{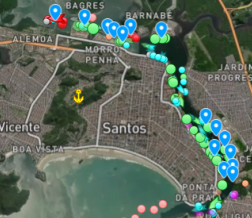}
\caption{Left: tides chart for the Rio de Janeiro (RJ) city, taken from the Tides Chart website. Right: vessel positions near the Santos (SP) port on a given day. Taken from the Marine Traffic website.}
\label{fig:tides}
\end{figure}

The experiments presented in this work were run with a corpus of Brazilian Portuguese verbalizations for the Blue Amazon domain. We initially developed web crawlers which extracted daily information from publicly available sources, including weather, temperature, tides charts, earthquakes, vessel positioning and oil extraction. Weather data and tides charts are extracted through the Tides Chart website, which provides information about high tides, low tides, tide charts, fishing times, ocean conditions, water temperatures and weather forecasts for thousands of cities around the world. Figure \ref{fig:tides} (left) shows an example of tides charts for the following week in Rio de Janeiro (RJ).

Vessel positioning is collected from the Marine Traffic website, which is an open, community-based project that provides real-time information about ship movements around the world and also their current location in ports and harbors. Figure \ref{fig:tides} (right) shows an example of vessel positions near the Santos (SP) port on a given day. Real time data regarding earthquakes in the Brazilian coast are taken from the Seismological Center at the University of São Paulo, and information regarding oil extractinon is obtained from the Brazilian government portal. After the data are collected and cleaned, they are stored in the MongoDB database, a NoSQL document-oriented database program which provides more flexibility and scalability over relational databases when input data is constantly changing \cite{stonebraker2010sql}.

We created the corpus based on information collected during 90 consecutive days for 50 cities in the Brazilian coast, and then performed content selection for past time-series data using feedback from domain experts. The intent messages were then sorted using a rule-based approach, and verbalizations of the intent messages were performed by the authors based on a sample of 300 texts extracted from the data. Syntactic and lexical patterns in the samples were used to produce a variety of target intent texts. Finally, intermediate representations in the pipeline steps were annotated in a intent-attribute-value format and used as input for the neural end-to-end approach. Some examples of intent-attribute-value in the dataset are:

\noindent
{\tt \footnotesize LOCATION(city="Santos",uf="SP",timestamp="Jan 15, 2022"); \\
WEATHER(condition="sunny",temperature="32ºC"); \\
EARTHQUAKE(magnitude="1.4 mR",depth="15km"); \\
VESSELS IN PORT(quantity="350",trend="high",days max="30")}

\section{Template Architecture}\label{sec:template}

Template-based data-to-text NLG systems directly translate non-linguistic input to linguistic surface structure by filling gaps in predefined template texts \cite{reiter1995nlg}. Because only the predefined variables can change in static templates, problems with maintainability and scalability arise from this approach; static template-based systems cannot be readily used to design sentence planning modules like discourse ordering, sentence formation and lexicalization. However, the main advantage of template-based approaches happens in cases where good linguistic rules are not yet available or in highly specific conditions where only a few texts are possible, and thus there is no real benefit in having a highly complex NLG system \cite{smiley2018e2e}.

For example, a simple template-based system might start out from a semantic representation saying that a new earthquake with a magnitude of 1.7 mR and depth of 10km was detected by the Seismology Center at the University of São Paulo in the city of Arapiraca, Alagoas (AL): \\
{\tt \footnotesize EARTHQUAKE(city="Arapiraca", uf="AL", magnitude="1.7mR", depth="10km", entity="Seismology Center at the University of São Paulo")}

This intent-attribute-value input is then directly associated with a template using a rule-based approach, such as: \\
{\tt \footnotesize A new earthquake was detected in [location] with a magnitude of [magnitude] and depth of [depth], by the [entity]. Stay safe!}

In this example, the gaps represented by [location], [magnitude], [depth] and [entity] are filled by looking up the relevant information in a table and the text will be generated only when a new earthquake is detected.

Some examples of template-based systems which publish daily reports on Twitter in Portuguese are Rosie from the Serenata de Amor operation, which identifies public funds expenses with discrepancies and indicates the reasons that lead it to believe they are suspicious, and Rui Barbot, which monitors stalled processes in the Supreme Federal Court of Brazil (STF) \cite{furtado2020automated}. 

\section{Pipeline Architecture}\label{sec:pipeline}

\begin{figure}[t]
\centering
\hspace*{-1cm}\includegraphics[width=0.94\textwidth]{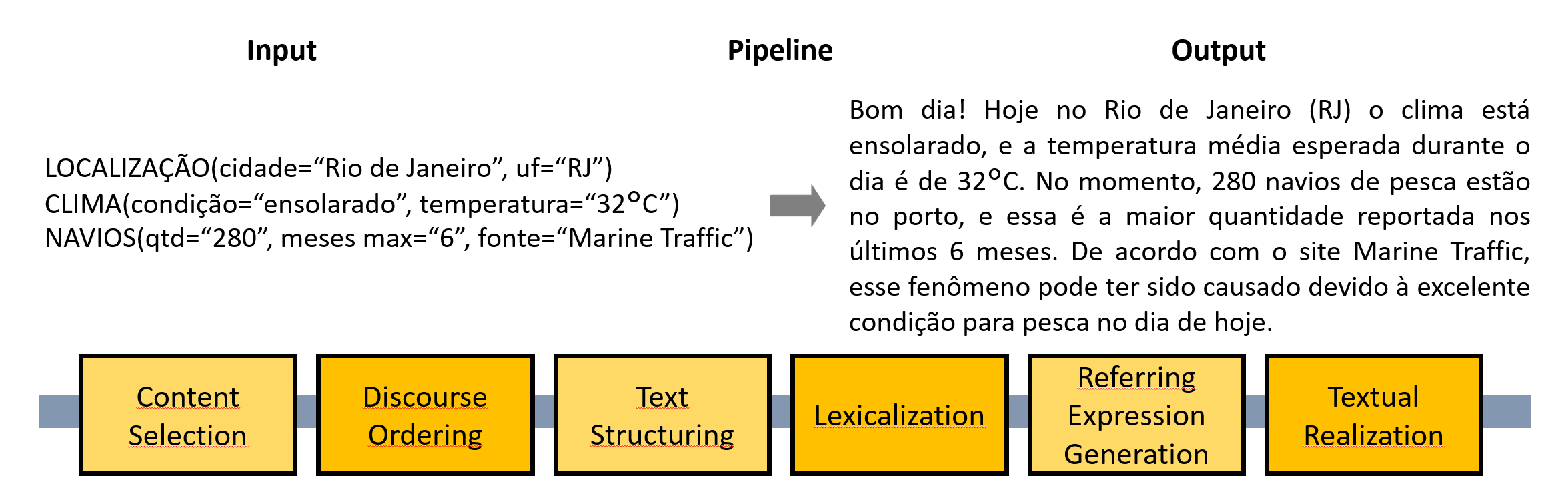}
\caption{Pipeline architecture with an example of a generated text for our automated journalism application.}
\label{fig:figura1}
\end{figure}

The pipeline architecture converts structured input data to output text in 6 sequential steps: \textit{Content Selection, Discourse Ordering, Text Structuring, Lexicalization, Referring Expression Generation and Textual Realization} \cite{horacek2001building}. Figure \ref{fig:figura1} shows the pipeline architecture steps with an example in the case of our automated journalism application. The system receives as input information regarding location, weather and vessels, and outputs the following text translated to English: 

\textit{\footnotesize Good Morning! Today in Rio de Janeiro (RJ) the weather is sunny, and the average temperature expected during the day is 32°C. Currently, 280 fishing vessels are in port, and this is the highest number of vessels reported in the last 6 months. According to the Marine Traffic website, this phenomenon may have been caused due to the excellent conditions for fishing today.
}

\textbf{Content Selection} In the first step of the pipeline architecture, the Content Selection module decides what information is important to be conveyed to the target audience and verbalized in the text. Since content determination precedes language generation, template-based systems can treat it in the exact same ways as the pipeline-base systems. While the former tend to take their departure from structured database records, the latter often use richer input, where some decisions concerning linguistics have already been made \cite{deemter2005real}. This step often requires the assistance of a domain expert to understand what information is relevant in the given context and how to group them in intent messages. The following text is an example of the content selection module output for our application:

\noindent
{\tt \footnotesize LOCATION(city="Santos",uf="SP",timestamp="Jan 15, 2022");  \\
\footnotesize WEATHER(condition="sunny",temperature="32ºC"); \\
\footnotesize VESSELS IN PORT (quantity="350",trend="high",days max="30"); \\
\footnotesize OCEAN(fishing condition="excellent",height of the sea:"1.8 meters");
}

\textbf{Discourse Ordering} Once the relevant content has been selected, the next step of the pipeline is to determine the order in which the intent messages should be verbalized to enhance reader comprehension \cite{heilbron2019tracking}. Although some authors have had success with machine learning  solutions to order facts for discourse planning \cite{dimitromanolaki2003learning}, most applications utilize a rule-based approach. For example, a possible outcome order might be: \\
{\small {\tt \footnotesize LOCATION, TEMPERATURE, EXCELLENT WEATHER AND FISHING CONDITIONS → CAUSES → PEAK OF VESSELS IN PORT, OIL EXTRACTION}}

\textbf{Text Structuring} Also referred to as Sentence Aggregation by some authors, Text Structuring is a NLG sub-task in which intents are organized into sentences and paragraphs. Given a linearized set of intent messages, the goal of this step is to generate predicates segmented by sentences. While it is possible to use a dedicated attention mechanism \cite{juraska2021attention}, most applications utilize explicit content text structuring.

For the case of our application, a possible text structure for the output of this module might be: 

\noindent
 {\tt \footnotesize  
Paragraph 1: LOCATION, TEMPERATURE \\
Paragraph 2: EXCELLENT WEATHER AND FISHING CONDITIONS → CAUSES → PEAK OF VESSELS IN PORT \\
Paragraph 3: OIL EXTRACTION}

\textbf{Lexicalization} The lexicalization step aims to find the proper word and phrases to express the content in each sentence. This is performed by adding words, phrases or word patterns to a language's vocabulary to inflect words based on their grammatical use (tense, number, case and gender, for example) and verbalize the intents \cite{stede1994lexicalization}. For example, we might have as intent-attribute-value input: \\
{\small {\tt \footnotesize LOCATION(city="Rio de Janeiro",uf="RJ",timestamp="Jan 15, 2022") \\
WEATHER(condition="cloudy",temperature="28ºC",max\_since\_days="10")}} \\
And generates the following output texts in Portuguese, and translated to English: \\
\textit{ \footnotesize Hoje, dia 15 de Janeiro de 2022, o clima é nublado no Rio de Janeiro (RJ). A temperatura média esperada é de 28ºC, e esta é a maior temperatura dos últimos 10 dias. \\
(Today, January 15th, 2022, the weather is cloudy in Rio de Janeiro (RJ). The average expected temperature is 28ºC, and this is the highest temperature in the last 10 days.)}

\textbf{Referring Expression Generation} In order to replace entity tags throughout the template, this module aims to automatically generate noun phrases to refer to entities mentioned as discourse unfolds. There are neural-based approaches that generate referring expressions for entities not found during the training process, but we  used a list of possible expressions for each entity. For the first reference to an entity in the text, a full description is used (e.g., {\tt \footnotesize WEBSITE → "Marine Traffic"}), whereas for subsequent references a random referring expression to the entity is chosen (e.g., {\tt \footnotesize WEBSITE → "the website; "the site"; "the Marine Traffic website"; "it"}; etc.).

\textbf{Textual Realization} The last step of the pipeline approach performs the remaining adjustments to transform intermediate machine representations into text. Detokenization, contractions, nominal and verbal are performed in order to make the content grammatically consistent. For  our robot-journalism application, this step applies a final layer of textual manipulation, to make the content more appealing to the target audience. The resulting texts are published every day using Twitter's API. 

An example of pipeline-based natural language application which publishes daily reports on Twitter in Portuguese is DaMata, a robot-journalist system covering the Brazilian Amazon deforestation \cite{teixeira2020damata}.

\section{End-to-End Architecture}\label{sec:endtoend}

End-to-end architectures for natural language generation recently gained popularity due to the massive amount of data and computational power available \cite{chen2014big}. Given enough labeled data, it does become possible go learn a mapping function which converts non-linguistic input into human-readable text without explicit use of intermediate representations. Such architectures operate by
applying deep neural networks, convolutional neural networks, recurrent neural networks (RNN) and transformers \cite{li2017deep}. Successful examples of end-to-end robot-journalism applications can be found in contexts where making mistakes and hallucinating content is not critical, for example data storytelling and image captions \cite{he2018deep}.

\begin{figure}[t]
\centering
\includegraphics[width=.98\textwidth]{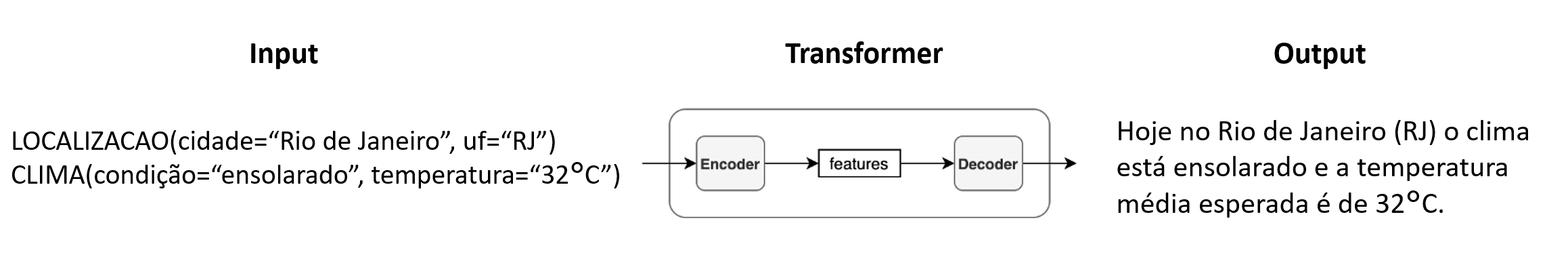}
\caption{Example of a generated text for our automated journalism application using transformers.}
\label{fig:figura2}
\end{figure}

Our goal here was to learn a direct mapping from a intent-attribute-value input text to a human-readable Portuguese text. We tested four   text-to-text transformer-based models: Bart \cite{lewis2019bart}, T5 \cite{xue2020mt5}, Blenderbot \cite{shuster2022blenderbot} and GPT2 \cite{radford2019language}. While GPT2 utilizes a decoder only module, the first three models use a encoder-decoder scheme \cite{cho2014properties}. Encoder-decoder networks have two distinct modules: the encoder, which transforms the input sequence into one feature vector, and a decoder which generates the output sequence. Figure \ref{fig:figura2} shows an example of this process in the context of our automated journalism application: the system receives the text sequence: {\tt \footnotesize Location(city= "Rio de Janeiro", state="RJ"); Weather(climate="sunny", temperature="32ºC")} and outputs the following text, translated to English: \textit{Today in Rio de Janeiro (RJ), the weather is sunny, and the average temperature expected during the day is 32ºC}.

For the experiments in this work, the dataset was randomly split into training (60\%), validation (20\%) and testing (20\%). The models were trained on Google Colab using a NVIDIA Tesla K80 GPU for a maximum of 100 epochs with a learning rate of 1e-5 and an early stopping criteria of 3 epochs.

\section{Comparing Architectures: Results}\label{sec:results}

The results of the three different automated journalism architectures compared in this work were evaluated by averaging the opinion of 5 domain experts in a sample of 100 texts present in the test set. For each pair of input data and output text, the participants rated the trials based on the fluency (i.e., \textit{``is the text easy to read?"}), semantics (i.e., \textit{``does the text clearly express the input data?"}) and lexical variety (i.e., \textit{``is the text original or is the content being repetitive?"}) of the Brazilian Portuguese output text in a 1-5 Likert scale \cite{joshi2015likert}, where 1 means strongly disagree and 5 means strongly agree. The neural end-to-end models were also evaluated using four different commonly used natural language generation metrics: Bleu, \cite{papineni2002bleu}, Gleu \cite{mutton2007gleu}, Rouge \cite{lin2004looking} and Meteor \cite{lavie2009meteor}. Table \ref{tab:results1} depicts the results of both automatic and human evaluations.

\begin{table}[t]
\caption{Comparing the automated journalism architectures on the Blue Amazon dataset.}
\label{tab:results1}
\vspace*{3ex}
\centering
\small 
\hspace*{-1mm}\begin{tabular}{l|l|l|l|l|l|l|l}
\hline
\textbf{Architecture} & \textbf{Bleu} & \textbf{Gleu} & \textbf{Rouge} & \textbf{Meteor} & \textbf{Fluency} & \textbf{Semantics} & \begin{tabular}{c} \textbf{Lexical} \\ \textbf{Variety}\end{tabular}
\\ \hline \hline

Template & - & - & - & - & \textbf{4.8} & \textbf{4.8} & 3.2\\ \hline

Pipeline & - & - & - & - & 4.5 & 4.6 & \textbf{4.4}\\ \hline

\begin{tabular}{l}End-to-end\\(Bart)\end{tabular} & \textbf{47.9} & \textbf{47.2} & 57.7 & 71.6 & 3.8 & 3.5 & 3.7\\ \hline

\begin{tabular}{l}End-to-end\\(T5)\end{tabular} & 43.6 & 46.5 & \textbf{58.2} & \textbf{73.2} & 3.5 & 3.3 & 3.7 \\ \hline

\begin{tabular}{l}End-to-end\\(Blenderbot)\end{tabular} & 38.4 & 36.1 & 52.9 & 60.3 & 3.2 & 3.2 & 3.1 \\ \hline

\begin{tabular}{l}End-to-end\\(GPT2)\end{tabular} & 19.3 & 18.4 & 17.8 & 19.8 & 2.1 & 1.9 & 2.2

\\\hline

\end{tabular}
\end{table}

\section{Discussion}\label{sec:discussion}

The results presented in Table \ref{tab:results1} show that the template-based architecture outperformed all the others in both fluency and semantics. As shown in the example of table \ref{tab:results2}, this happens because the output texts are very repetitive and the architecture fails to provide lexical variety. This is a known limitation of the approach \cite{pereira2015towards}, and after the results obtained in this work we opted to apply the template-based architecture only for sensitive and edge case scenarios, where communicating the message on a very fluent and objective way is more important than having well connected sentences and lexical variety. Examples of these scenarios in the Blue Amazon domain are, for example, when a new earthquake is detected or when oil extraction reaches a critical level.

Although the pipeline-based architecture presented less fluency and semantics than the template-based approach, it also obtained high scores in these metrics and received the best overall score for lexical variety. This means that in domains where there are enough linguistic insights and computational resources available to develop a pipeline-based architecture, and also there is no critical or sensitive information to be conveyed, the pipeline-based architecture is more interesting than the template-based approach because it provides for more lexical variety to the target audience. It also has the advantage of not hallucinating data, unlike the novel end-to-end approaches.

Results from the neural end-to-end architecture show that all the four tested transformer-based models scored significantly less in all the human quantitative metrics. This happens because deep learning methods for NLG often hallucinates data and does not convey all the meaningful information in the input, as shown in table \ref{tab:results2}. This approach also makes more lexical and semantic mistakes compared to the pipeline-based architecture, given that the latter has a dedicated lexical module while the former does not. The main advantage in this approach is in domains where hallucinating data is not critical, there are no domain experts to provide linguistic insights, which is not the case for our application.

As for the automatic metrics, the Bart model outperformed the other neural end-to-end models in both Bleu and Gleu, while the T5 achieved the best Rouge and Meteor scores. While Blenderbot obtained average scores, GPT2 obtained the lowest scores overall for all the purposed metrics. Works in simmilar contexts have reported BLEU scores ranging from 45\% to 65\% \cite{duvsek2018findings}. Table \ref{tab:results2} show some examples of input and output pairs for each of the compared architectures. The template-based approach outputs texts on a very objective and straightforward way, while the pipeline-based approach outputs more complex text with connected sentences and more lexical variety. The neural-based approach sometimes hallucinates both data and content, and also makes both lexical and semantic mistakes. Unlike the other 3 models, GPT2 utilizes a decoder only module, and the result was that it failed to complete long sentences and properly transform input data into coherent output text.

\begin{table}[hbt!]
\footnotesize
\caption{Examples of input and output pairs for some of the compared architectures for automated journalism.}
\label{tab:results2}
\centering
 \begin{adjustwidth}{-0.7cm}{}
\begin{tabular}{|p{1.8cm}|p{5cm}|p{3.8cm}|p{4.45cm}|}
\hline
\textbf{Architecture} & \textbf{Input} & \textbf{Output} & \textbf{Conclusion} \\ \hline

Template &
{\tt LOCALIZAÇÃO(cidade= "Santos", uf="SP"); 
CLIMA(condição="nublado", temperatura="26ºC", vento="18km/h"); 
TERREMOTO(magnitude="1.3 mR", depth="10km")} & 
\textit{Hoje em Santos (SP) a previsão é de tempo nublado. A temperatura é de 26ºC. O vento é de 18km/h. Foi detectado um terremoto de magnitude 1.3 mR e profundidade de 10km.} &
 The architecture managed to accurately convey all the information present in the input data without any lexical or semantic mistake. However, the text is too simple, lacks connection between sentences and does not provide for lexical variety.
\\ \hline

Pipeline &
{\tt LOCALIZAÇÃO(cidade= "Cabo Frio", uf="RJ"); CLIMA(condição= "ensolarado", temperatura="33ºC", mar="0,8m", umidade="76\%", nebulosidade="54\%", vento="29km/h", protetor="sim")} & 
\textit{Hoje a previsão é de tempo ensolarado, com uma temperatura média esperada de 34ºC. A umidade de 54\%, a nebulosidade de 29km/h e a velocidade do vento de 54\%. Utilize protetor solar se for sair de casa!} &
 The architecture managed to accurately convey all the information present in the input data without any lexical or semantic mistake, and also provided for well connected sentences.
\\ \hline

End-to-end (Bart) &
{\tt LOCALIZAÇÃO(cidade= "Itajaí", uf="PE"); CLIMA(condição= "ensolarado", temperatura="25ºC"); NAVIOS(quantidade="185", dias max="28")} & 
\textit{Hoje em Itajaí (PE) foi detectada uma quantidade de 285 navios no porto da cidade, essa e esse é o maior valor registrado nos últimos 28 dias.} &
The architecture output a text with correct semantic and fluency, but forgot to convey weather related data and made a mistake in the amount of vessels in the port.\\ \hline

End-to-end (T5) &
{\tt LOCALIZAÇÃO(cidade= "Itajaí", uf="PE"); NAVIOS(quantidade="180", temperatura="35ºC", mar="0,8m", umidade="76\%", vento="29km/h", protetor solar="sim")} & 
\textit{Hoje a previsão é de tempo temperatura média esperada de 35ºC a umidade é de 29km/h a velocidade do vento é de 76\%. Utilize protetor solar!} &
 The architecture output a text with some lexical and fluency mistakes, failing to properly connected sentences. It also swapped the values of humidity and wind speed, which would be critical on a more sensitive scenario.
\\ \hline

End-to-end (GPT2) &
{\tt LOCALIZAÇÃO(cidade= "Recife", uf="PE"); CLIMA(condição= "ensolarado", temperatura="25ºC", mar="1,8m", umidade="58\%, vento="21km/h")} & 
\textit{Hoje aaa temperatura é de 25°V, aaa temperatura é de 25°V, aaa temperatura é de 25°V.} &
 The architeture swapped °C with °V for no particular reason and failed to convey most of the information in the input data. It also repeated the temperature value three times and output nonexistent words.
\\\hline

\end{tabular}
 \end{adjustwidth}
\end{table}

\begin{figure}[t]
\centering
\includegraphics[width=.85\textwidth]{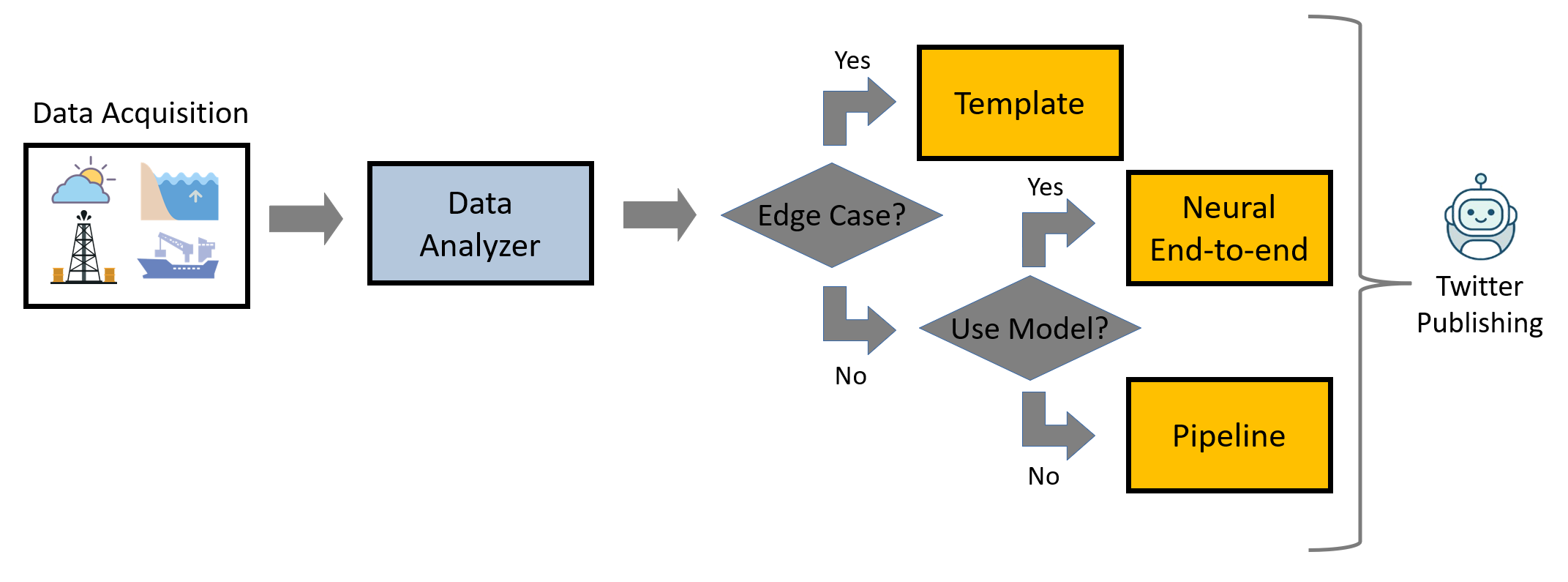}
\vspace*{-2ex}
\caption{Automated journalism system architecture developed to publish daily reports about the Blue Amazon.}
\label{fig:figura3}
\end{figure}

\section{Conclusion}\label{sec:conclusion}

This paper compared data-to-text natural language generation architectures in the context of automated journalism for the Blue Amazon domain. We also created a publicly available dataset and an automated journalism system which publishes daily Brazilian Portuguese reports on Twitter by collecting, storing and analyzing information from multiple sources. Figure \ref{fig:figura3} shows the current architecture of our automated journalism application. After the data is collected and stored on our system's database,  a rule-based data analyzer module  decides whether the content is critical. If that is the case, we opted to use the Template approach  to communicate the message and to avoid the risk of generating texts with poor fluency or semantics. If the data analyzer module decides that the information is not critical, results show  that the pipeline approach produces better texts than    neural end-to-end approaches, by avoiding data hallucination while also generalizing better to unseen inputs. For our application, we decided to implement a Boolean input parameter responsible for choosing between the pipeline and the neural end-to-end approaches. We have scheduled a weekly retraining of the neural end-to-end models to make sure that they will continue to perform as expected for new inputs, thus avoiding the risk of data drift.

In the future, we plan to add more sources of information to our application, such as real time reporting of illegal fishing activities and other natural disasters. We also plan to use different neural end-to-end methods at each step of the pipeline architecture, and to replace numbers with their equivalent text to validate whether better overall results can be achieved. We  plan to  experiment  with text summarization architectures  to outline public news related to the Blue Amazon in short tweets.

\section{Acknowledgements}\label{sec:acknowledgements}

We would like to thank the Center for Artificial Intelligence (C4AI: www.c4ai.inova.usp), supported by  São Paulo Research Foundation (FAPESP grant \#2019/07665-4) and   IBM Corporation. This research was also partially supported by the Coordenação de Aperfeiçoamento de Pessoal de Nível Superior (CAPES), Finance Code 001. M.\ M.\ José, has been supported by the Itaú Scholarship Program (PBI), run by the Data Science Center (C2D) of the Escola Politécnica da Universidade de São Paulo. F.\ G.\ Cozman is partially supported by the National Council for Scientific and Technological Development of Brazil (CNPq) grant \#312180/2018-7. Any opinions, findings, and conclusions expressed in this manuscript are those of the authors and do not necessarily reflect the views, official policy, or position of the financiers.

\bibliographystyle{sbc}
\bibliography{sbc-template}

\end{document}